\documentclass[runningheads]{llncs}

\usepackage{eccv}

\usepackage{eccvabbrv}

\usepackage{graphicx}
\usepackage{booktabs}
\usepackage{multirow}

\usepackage[accsupp]{axessibility}  

\usepackage[pagebackref,breaklinks,colorlinks,citecolor=eccvblue]{hyperref}

\usepackage{orcidlink}

\usepackage{bm}
\usepackage[symbol]{footmisc}

\begin{document}

\title{ByteEdit: Boost, Comply and Accelerate \\ Generative Image Editing} 

\titlerunning{ByteEdit}

\author{
Yuxi Ren\inst{*} \and Jie Wu\inst{* \dagger} \and Yanzuo Lu\inst{*} \and Huafeng Kuang \and
Xin Xia \and \\ 
Xionghui Wang \and Qianqian Wang \and Yixing Zhu \and
Pan Xie \and Shiyin Wang \and \\
Xuefeng Xiao \and Yitong Wang \and Min Zheng \and Lean Fu
}

\authorrunning{Y.~Ren et al.}

\institute{
ByteDance Inc.\\
Project Page: \url{https://byte-edit.github.io}
}

\maketitle
\footnotetext[1]{Equal contribution.}
\footnotetext[2]{Corresponding author.}

\begin{figure}
    \centering
    \vspace{-0.8cm}
    \includegraphics[width=0.9\linewidth]{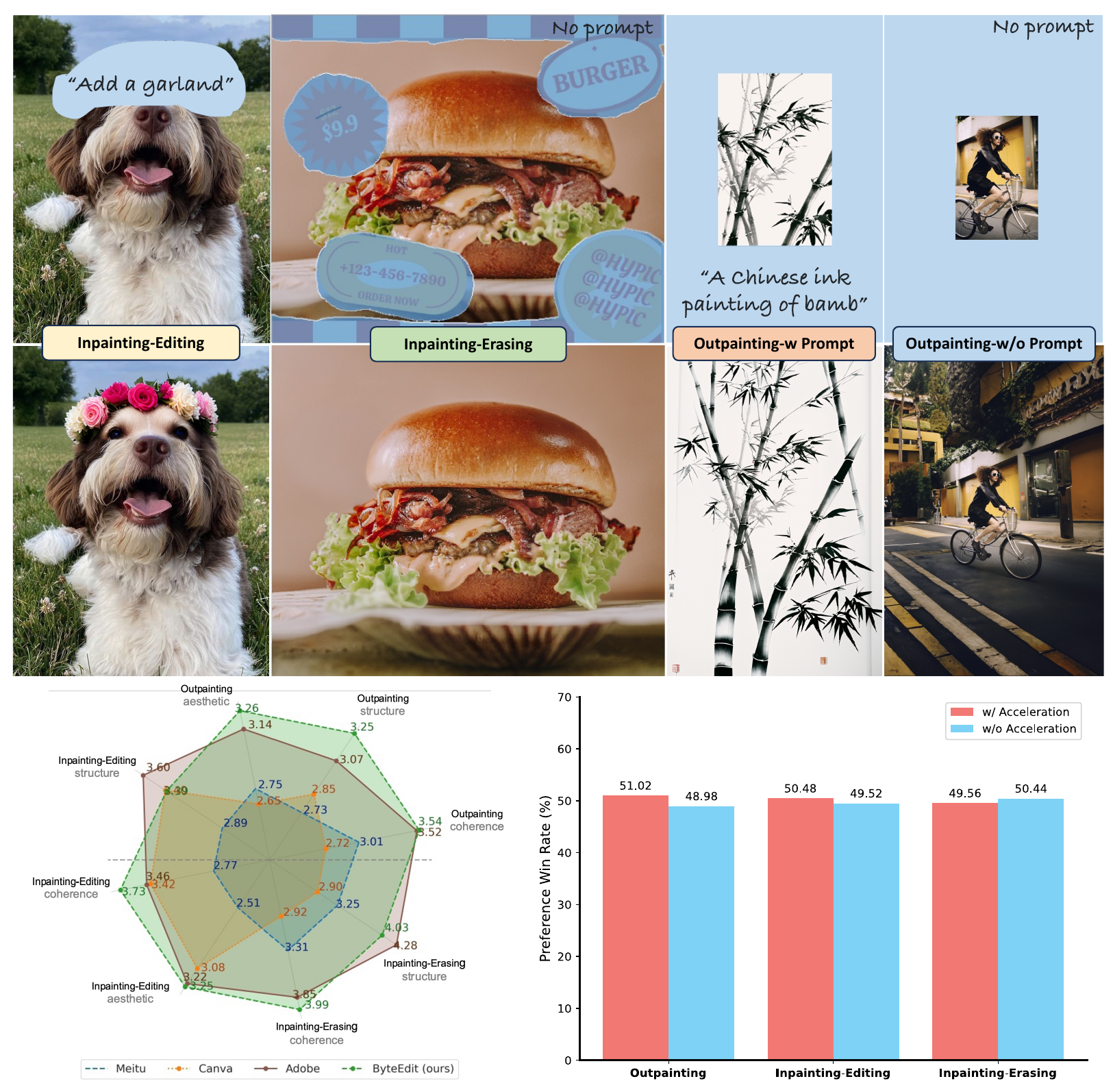}
    \caption{We introduce \textbf{\textit{ByteEdit}}, a novel framework that utilizes feedback learning to enhance generative image editing tasks, resulting in outstanding generation performance, improved consistency, enhanced instruction adherence, and accelerated generation speed. To the best of our knowledge, ByteEdit emerges as the most \textit{superior} and the \textit{fastest} solution currently in the field of generative editing.}
    \label{fig:teaser}
\end{figure}

\begin{abstract}
Recent advancements in diffusion-based generative image editing have sparked a profound revolution, reshaping the landscape of image outpainting and inpainting tasks. Despite these strides, the field grapples with inherent challenges, including: i) inferior quality; ii) poor consistency; iii) insufficient instrcution adherence; iv) suboptimal generation efficiency.
To address these obstacles, we present \textit{\textbf{ByteEdit}}, an innovative feedback learning framework meticulously designed to \textit{\textbf{B}}oost, Compl\textit{\textbf{y}}, and Accelera\textit{\textbf{te}} Generative Image \textit{\textbf{Edit}}ing tasks. ByteEdit seamlessly integrates image reward models dedicated to enhancing aesthetics and image-text alignment, while also introducing a dense, pixel-level reward model tailored to foster coherence in the output. Furthermore, we propose a pioneering adversarial and progressive feedback learning strategy to expedite the model's inference speed.
Through extensive large-scale user evaluations, we demonstrate that ByteEdit surpasses leading generative image editing products, including Adobe, Canva, and MeiTu, in both generation quality and consistency. ByteEdit-Outpainting exhibits a remarkable enhancement of \textbf{388\%} and \textbf{135\%} in quality and consistency, respectively, when compared to the baseline model.
Experiments also verfied that our acceleration models maintains excellent performance results in terms of quality and consistency.

\end{abstract}

\section{Introduction}
\label{sec:intro}
The field of generative image editing has experienced remarkable advancements in recent years~\cite{wang2023imagen,xie2023smartbrush,lugmayr2022repaint,yang2023paint,avrahami2022blended,yu2023inpaint,xie2023dreaminpainter,yildirim2023inst,nichol2021glide,chen2023anydoor,lu2024coarse}, propelled by the development of diffusion models~\cite{ho2020denoising,rombach2022high,qin2024diffusiongpt,zhang2023diffusionengine,ren2023ugc}. 
This progress has led to the emergence of influential products that have reshaped the landscape of image editing. A notable example is Adobe Firefly~\cite{adobe}, which has revolutionized the creative process by enabling users to seamlessly incorporate, extend, or remove content from images through simple text prompts, thereby transcending the traditional boundaries of Photoshop.
In our paper, we focus on the domain of generative image editing, with particular emphasis on two key aspects: 1) \textit{Outpainting}, involving the expansion of the surrounding scene in an image based on provided input or even without explicit prompts, and 2) \textit{Inpainting}, encompassing the random masking of specific image regions followed by the generation of corresponding content guided by given prompts (Inpainting-Editing) or erase certain objects (Inpainting-Erasing).
Despite the notable advancements achieved through diffusion-based algorithms, several challenges persist within this field:

\noindent \textit{\textbf{Inferior Quality}}: the quality of generated images frequently falls short in terms of realism, aesthetic appeal, and fidelity to minute details.

\noindent \textit{\textbf{Insufficient Instruction Adherence}}: The existing models grapple with the arduous task of faithfully adhering to provided instructions, resulting in a lack of alignment between the generated image and the input text;

\noindent \textit{\textbf{Poor Consistency}}: The generated regions exhibit an unsatisfactory level of coherence with the original image, manifesting as a deficiency in terms of color, style, texture, and other salient visual attributes;

\noindent \textit{\textbf{Suboptimal Generation Efficiency}}: The generation process is characterized by sluggish speeds and inadequate efficiency, thereby imposing significant obstacles when confronted with large-scale image editing endeavors.

Recently, various efforts have been made to address the aforementioned challenges in the field. For instance, Imagen Editor~\cite{wang2023imagen} has employed an object detection approach to extract inpainting masks, while simultaneously capitalizing on original high-resolution images to faithfully capture intricate details.  SmartBrush~\cite{xie2023smartbrush} has adopted a multi-task training strategy coupled with precision controls, encompassing both text and shape guidance, to enhance visual quality, mask controllability, and preserve the background. Additionally, RePaint~\cite{lugmayr2022repaint} has utilized a pre-trained unconditional DDPM~\cite{ho2020denoising} prior and ingeniously modified the reverse diffusion iterations to generate high-quality and diverse output images.
However, these approaches have primarily focused on addressing singular problems and have yet to achieve a more comprehensive solution. Large Language Models (LLMs) has made a notable surge in incorporating learning based on human feedback, and initial endeavors have been undertaken in the Text-to-Image (T2I) domain~\cite{dong2023raft,lee2023aligning,xu2023imagereward,zhang2024confronting,yuan2024self,yang2024dense}. Inspired by these developments, we pose the question: \textit{Can we leverage human feedback to guide generative image editing to unleash the potential for superior generation outcomes?}

This paper introduces ByteEdit, an innovative framework for optimizing generative image editing through the incorporation of feedback learning.
ByteEdit builds multiple reward models, namely the Aesthetic reward model, Alignment reward model, and Coherent reward model, to achieve exceptional generation effects, improved instruction adherence and enhanced consistency,  respectively. 
These carefully designed reward models serve as the foundation for our proposed perceptual feedback learning (PeFL) mechanism, which provides task-aware and comprehensive supervision signals.
Moreover, ByteEdit introduce an adversarial feedback learning strategy that employs the trainable reward model as the discriminator. This strategy ensures that the model benefits from the PeFL supervision and provide clear images even during high-noise stages, further improves both the performance and speed of our model.
To expedite the sampling process, a progressive training strategy is empolyed to gradually reduce the optimization time steps and facilitate model inference in a low-steps regime.

\begin{itemize}

\item ~\underline{\textit{New Insight}}: 
To the best of our knowledge, we offer the first attempt to incorporate human feedback into the field of generative image editing. ByteEdit significantly enhances the overall performance of the model across various key aspects, opening new horizons in this field of study.

\item ~\underline{\textit{Comprehensive Framework}}: 
By designing complementary global-level and pixel-level reward models, we effectively guide the model towards achieving improved beauty, enhanced consistency, and superior image-text alignment. 

\item ~\underline{\textit{Efficiency and Pioneering}}: 
Progressive feedback and adversarial learning techniques are introduced to accomplish a remarkable acceleration in the model's inference speed, all while maintaining a minimal compromise on output quality. Notably, ByteEdit stands as the first successful endeavor in accelerating generative editing models.

\item ~\underline{\textit{Outstanding Performance}}: 
Extensive user studies show that ByteEdit exhibits obvious advantages in terms of quality, consistency, efficiency, and speed, compared to the most competitive products. ByteEdit emerges as the fastest and most superior solution currently available in image editing.

\end{itemize}

\section{Related Work}

\noindent\textbf{Generative Image Editing.}
Generative Image Editing is a research area focused on filling user-specified regions of interest with desired contents. 
GLIDE ~\cite{nichol2021glide} is the pioneering work that introduced text-to-image diffusion for editing purposes, and Repaint ~\cite{lugmayr2022repaint}, on the other hand, conditions an unconditionally trained model (e.g. DDPM~ \cite{ho2020denoising}) and leverages visible pixels to fill in missing areas.
To enable more precise editing control, Blended Diffusion ~\cite{avrahami2022blended} incorporates multimodal embeddings and enforces similarity between images and text using CLIP ~\cite{radford2021learning}. 
SmartBrush ~\cite{xie2023smartbrush} pushes the boundaries of mask generation by utilizing instance and panoptic segmentation datasets instead of random generation. 
Further improvements include the introduction of the powerful Segment Anything (SAM) ~\cite{kirillov2023segment} model by \cite{yu2023inpaint}, which achieves mask-free removal, filling, and replacing of multiple pipelines. 
Inst-Inpaint ~\cite{yildirim2023inst} specializes in text-driven object removal without the need for explicit masking. Additionally, this method proposes the GQA-Inpaint dataset, which comprises image pairs with and without the object of interest, facilitating effective object removal guided by textual input.
In addition to comparing our proposed method with existing academic approaches, we also benchmark against industry methods like Adobe~\cite{adobe}, Canva~\cite{canva}, and MeiTu~\cite{meitu}, providing a comprehensive evaluation across different domains and highlighting the strengths of our approach.

\noindent\textbf{Human Feedback Learning.}
Foundation models for text-to-image diffusion often rely on pre-training with large-scale web datasets, such as LAION-5B ~\cite{schuhmann2022laion}, which may result in generated content that deviates from human ethical values and legal compliance requirements. Previous approaches ~\cite{lee2023aligning,dong2023raft}  attempted to address this issue by constructing preference datasets using hand-crafted prompts or expert generators. However, these methods suffered from over-fitting due to their limited real-world scenarios and generator capabilities.
To overcome these limitations, researchers proposed various reward models trained with expert annotations ~\cite{wu2023human,xu2023imagereward}  or feedback from web users ~\cite{kirstain2024pick,isajanyan2024social} to enforce alignment with human preferences. Drawing inspiration from reinforcement learning with human feedback (RLHF) utilized in natural language processing (NLP), researchers explored the application of human feedback learning in text-to-image diffusion ~\cite{xu2023imagereward,yuan2024self,yang2024dense,zhang2024confronting} to achieve more realistic, faithful, and ethical outcomes.
Among these efforts, ImageReward  ~\cite{xu2023imagereward} primarily focused on overall image quality and overlooked the complexity of human perception.
In our work, we extend the concept of human feedback learning by introducing three fine-grained independent reward models tailored for generative image editing: aesthetics, image-text alignment, and pixel-level coherence.

\section{ByteEdit: Boost, Comply and Accelerate}



ByteEdit, focuses on generative image editing tasks that enable users to manipulate image content within a specific region of interest using textual descriptions. With an input image $x$, a region-of-interest mask $m$, and a user-provided textual description $c$, our primary objective is to generate an output image $y$ that preserves the unmasked region in the input image $x$, while aligning the masked region well with both the description of $c$ and visual attributes in $x$.
In this study, we introduce two key functionalities within ByteEdit: ByteEdit-Outpainting and ByteEdit-Inpainting. ByteEdit-Outpainting extends the image by generating content beyond the boundaries of the input image, while ByteEdit-Inpainting fills or erases in arbitrary areas of the input image.

\begin{figure}[t]
    \centering
    \includegraphics[width=\linewidth]{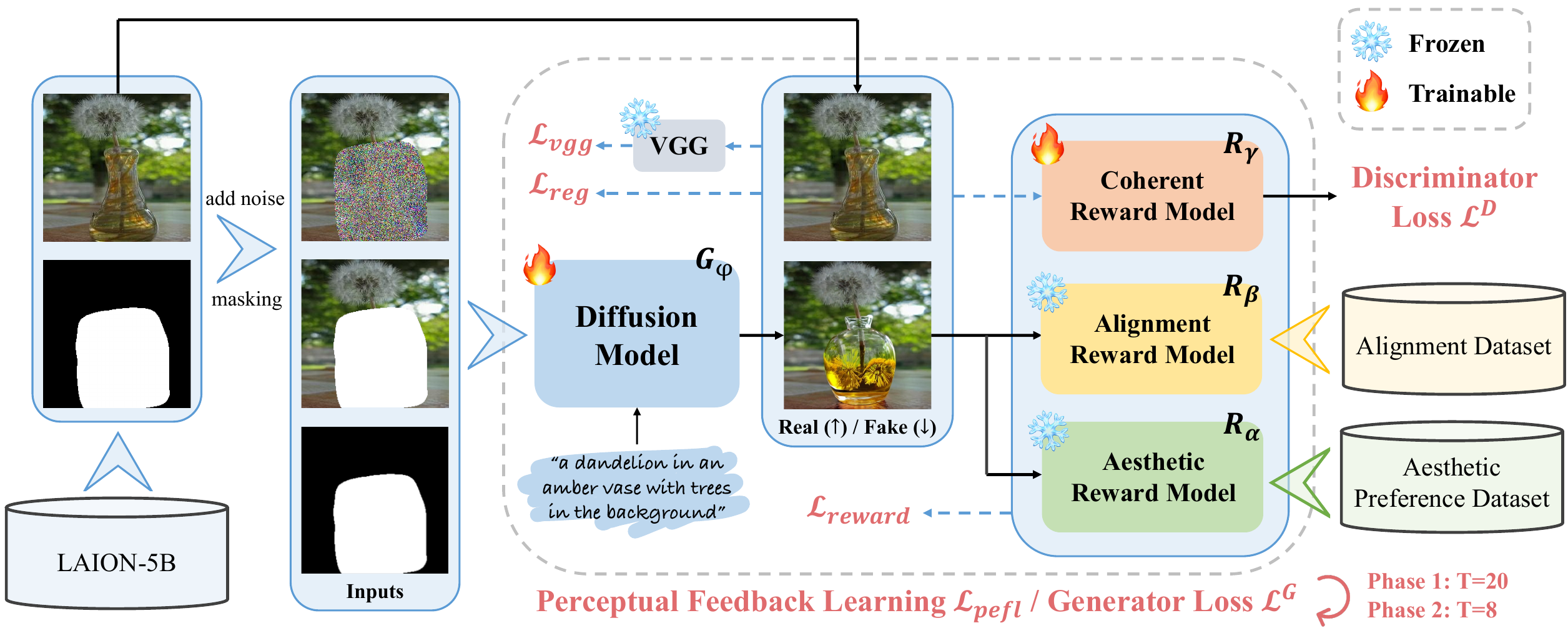}
    \caption{
        ByteEdit formulates a comprehensive feedback learning framework that facilitating aesthetics, image-text matching, consistency and inference speed.
    }
    \label{fig:pipeline}
\end{figure}

The ByteEdit pipeline is presented in Fig \ref{fig:pipeline}, providing an overview of the system's workflow. In the subsequent subsections, we delve into the details of two crucial components: Boost (\cref{sec:boost}) and Comply (\cref{sec:comply}). Furthermore, we elaborate on the Accelerate scheme in \cref{sec:accelerate}, which illustrates an approach to expedite the processing time and improve the efficiency of the ByteEdit system. 

\subsection{Boost: Perceptual Feedback Learning}
\label{sec:boost}
In the field of generative image editing, the persistent challenge of subpar quality has impelled us to propose a pioneering approach that introduces human feedback, hitherto unexplored in this domain. Our novel pipeline comprises three key components: feedback data collection, reward model training, and perceptual feedback learning.


\noindent\textbf{Feedback Data Collection.}
We first randomly extract more than 1,500,000 text prompts from the Midjourney Discord~\cite{turc2022midjourney} and MS-COCO Caption~\cite{chen2015microsoft} datasets.
To ensure the diversity, a clustering algorithm, namely K-Means, was employed, leveraging the similarities derived from state-of-the-art large language models~\cite{liu2023visual}. Further, the features were visualized in lower dimensions using t-SNE~\cite{van2008visualizing}, enabling the identification of data points possessing the largest average distance from their k-nearest neighbors.
We also manually eliminate less informative and decorative-dominanted prompts such as ``unbelivable'', ``fantastic'' and ``brilliant'' to improve the prompt quality.
This meticulous procedure yielded approximately 400,000 candidate prompts, exhibiting diverse distributions, which were subsequently subjected to multiple text-to-image diffusion models, including SD1.5~\cite{rombach2022high} and SDXL~\cite{podell2023sdxl}.
Those images which excessively inferior quality or ambiguous characteristic are manually removed.
Accompanying each prompt, a set of four generated images was presented to experts, who were tasked with selecting the best and worst images based on various aspects, encompassing aesthetic appeal, color composition, structural coherence, and brightness.
The resulting dataset, herein referred to as the aesthetic preference dataset $\mathcal{D}_{aes}$, encompasses a collection of massive triplets $(c,x_p,x_n)$, where $x_p$ and $x_n$ correspond to the preferred and non-preferred generated images of prompt $c$, respectively.

\noindent\textbf{Reward Model Training.}
Building upon this dataset, we follow the training techniques in \cite{xu2023imagereward} to learn an aesthetic reward model $R_{\alpha}(\cdot)$ of trainable parameters $\alpha$, which we briefly summarize here.
The image and text features of the input are extracted from the BLIP~\cite{li2022blip} backbone, combined with cross attention, and fed into an MLP to obtain an aesthetic score.
The training objective can be formulated as,
\begin{equation}\label{eq:aesthetic}
    \mathcal{L}(\alpha)=-\mathbb{E}_{(c,x_p,x_n)\sim\mathcal{D}_{aes}}[\log\sigma(R_{\alpha}(c,x_p)-R_{\alpha}(c,x_n))],
\end{equation}
where $\sigma(\cdot)$ represents the Sigmoid function used for normalization.

\noindent\textbf{Perceptual Feedback Learning.}
Leveraging the power of our crafted reward model, we specifically introduce Perceptual Feedback Learning (PeFL) to fine-tune diffusion models with human feedback for generative image editing.
Departing from the conventional practice of sequentially refining the predictions from the final step $x_T$ to the initial step $x_0^\prime$ ( $x_T \rightarrow x_{T-1} \rightarrow \cdots \rightarrow x_0^\prime$), we adopt an innovative perspective by performing optimization on the direct prediction outcomes $x_t \rightarrow x_0^\prime$ at various intermediate steps $t\in[1,T]$ ($T=20$ in this case).
Through this dynamic exploration across different stages of denoising, we uncover the following distinctive observations:
\begin{itemize}
    \item During the initial stages of denoising ($t\in[11,20]$), the generative model (i.e. U-Net~\cite{ronneberger2015u}) struggles to seamlessly complement the full image. Consequently, the reward model encounters challenges in accurately evaluating images that are hindered by obstacle masking.
    \item As the denoising process progresses ($t\in[1,10]$), the reward model is able to identify and reward visually appealing images even in the presence of mild noise.
\end{itemize}

Drawing upon these insightful observations, we present an innovative stage-wise approach, to amplify the potential of generative image editing. Our proposed method encompasses the following key stages:
1) In stage 1 ($t\in[16,20]$), we simply skip these steps with extremely heavy noise by diffusing the masked input $x \odot (1-m)$ into noisy latents at a fixed step $T_1=15$. This strategy is motivated by the fact that the generative model's ability to fill in intricate details within the masked region is limited in this timestep, thus rendering the training overhead unnecessary. This is the central difference between our approach and the ReFL proposed in \cite{xu2023imagereward}, whose training paradigm relies solely on prompts and starts inference from pure noise.
We aim to bring about a more pronounced correlation between the output and input image, thus facilitating the preservation of details in the unmasked areas;
2) In stage 2 ($t\in[t^\prime,15]$), we randomly select a denosing step $t^\prime\sim[1,T_2]$ ($T_2=10$ in this case) and perform inference without gradient starting from the noisy latent, i.e. $x_{T_1} \rightarrow x_{T_1-1} \rightarrow \cdots \rightarrow x_{t^\prime}$.
This method ensures that the complemented image generated at this stage exhibits a sufficiently high level of quality, rendering it more amenable to evaluation by the reward model;
3) In stage 3 ($x_{t^\prime} \rightarrow x_0^\prime$), the final stage entails the direct prediction of the complemented image $x_0^\prime$. We leverage the aesthetic score obtained from the reward model as invaluable human feedback to refine the generative model $G_\phi(\cdot)$. This refinement process is achieved through the utilization of the following loss function:
\begin{equation}\label{eq:reward}
    \mathcal{L}_{reward}(\phi)=-\mathbb{E}_{(x,m,c)\sim\mathcal{D}_{train},t^\prime\sim[1,T_2]}[\log \sigma(R_{\alpha}(c, G_\phi(x,m,c,t^\prime)))],
\end{equation}
where $\mathcal{D}_{train}$ represents the fine-tuning dataset (i.e. LAION-5B~\cite{schuhmann2022laion}).
The term $G\phi(x,m,c,t^\prime)$ denotes the decoded output image $x_0^\prime$ generated by the generative model at step $t^\prime$, given the masked input $x\odot (1-m)$ and the prompt $c$.
To further maintain the consistency and detail fidelity of the generated area and the original image area,we introduce pixel-level regularization  (i.e., L1 loss) and a perceptual loss, which captures the discrepancy in VGG features~\cite{simonyan2014very}. Collectively, these regularization techniques can be formulated as follows:
\begin{equation}\label{eq:l1}\begin{aligned}
    & \mathcal{L}_{reg}(\phi)=\mathbb{E}_{(x,m,c)\sim\mathcal{D}_{train},t^\prime\sim[1,T_2]}\|x-G_\phi(x,m,c,t^\prime)\|_1, \\
    & \mathcal{L}_{vgg}(\phi)=\mathbb{E}_{(x,m,c)\sim\mathcal{D}_{train},t^\prime\sim[1,T_2]}\|V(x)-V(G_\phi(x,m,c,t^\prime))\|_1,
\end{aligned}\end{equation}
where $V(\cdot)$ represents the VGG network.
The overall training objective of our PeFL can be summarized as,
\begin{equation}\label{eq:pefl}
    \mathcal{L}_{pefl}(\phi)=\mathcal{L}_{reward}+\eta(\mathcal{L}_{reg}+\mathcal{L}_{vgg}),
\end{equation}
where $\eta$ is a hyperparameter for balancing loss weights.

\subsection{Comply: Image-Text Alignment with Coherence}
\label{sec:comply}

Diverging from the text-to-image synthesis focus of \cite{xu2023imagereward}, our method encompasses an additional emphasis on assessing the \textbf{alignment} between the generated content of the masked area and the user-specified prompt, as well as ensuring \textbf{coherence} with the unmasked region at the pixel level. To achieve this, we introduce two further components in this section, which complement the aesthetic reward model $R_\alpha$ proposed earlier.

\noindent\textbf{Image-Text Alignment.}
We note that the presence of numerous poorly matched image-text pairs within the LAION dataset ~\cite{schuhmann2022laion}. Exploiting these pairs as non-preferred samples for training reward models allows us to reduce manual annotation costs significantly. Initially, we employ the CLIP model~\cite{radford2021learning} to identify image-text pairs with lower CLIPScore~\cite{hessel2021clipscore} from the LAION dataset. Subsequently, we leverage advanced multi-modal large language models such as LLAVA~\cite{liu2023visual} to generate more informative and descriptive captions for the input images. These captions are considered more accurate than the original prompts. This process yields approximately 40,000 triplets $(c_p,c_n,x)$ as alignment dataset $\mathcal{D}_{align}$, where $c_p$ and $c_n$ correspond to the preferred and non-preferred textual descriptions of the image $x$, respectively. These triplets are utilized for training the image-text alignment reward model, denoted as $R_\beta(\cdot)$. The architecture of $R_\beta$ mirrors that of $R_\alpha$, while the training objective is similar to Eq. \ref{eq:aesthetic}:
\begin{equation}\label{eq:alignment}
    \mathcal{L}(\beta)=-\mathbb{E}_{(c_p,c_n,x)\sim\mathcal{D}_{align}}[\log\sigma(R_{\beta}(c_p,x)-R_{\beta}(c_n,x))],
\end{equation}

\noindent\textbf{Pixel-Level Coherence.}
The issue of coherence arises from the presence of inconsistent content within and outside the regions of interest, characterized by subtle visual cues such as color variations, stylistic discrepancies, and textural differences. To tackle this challenge, a coherent reward model, denoted as $R_\gamma(\cdot)$,  is specifically designed for pixel-level discrimination, as opposed to the holistic evaluation performed by $R_\alpha(\cdot)$ and $R_\beta(\cdot)$. Our approach entails training a ViT-based~\cite{dosovitskiy2020image} backbone network, followed by a prediction MLP head, to assess the authenticity and assign a score to each pixel in the input image. By formulating the loss function as follows:
\begin{equation}\label{eq:coherent}
    \mathcal{L}(\gamma)=-\mathbb{E}_{
    (x,m,c)\sim\mathcal{D}_{train}, t^\prime\sim[1,T_2]
    }[\log\sigma(R_\gamma(z))+\log(1-\sigma(R_\gamma(z^\prime)))],
\end{equation}
where $z\sim x\in\mathbb{R}^{H\times W\times 3}, z^\prime\sim G_\phi(x,m,c,t^\prime)\in\mathbb{R}^{H\times W\times 3}$ are pixels of the corresponding image and $H, W$ represent the height and weight respectively.

\subsection{Accelerate: Adversarial and Progressive Training}
\label{sec:accelerate}

\noindent\textbf{Adversarial training.}
Concurrent works such as UFOGen~\cite{xu2023ufogen} and SDXL-Turbo~\cite{sauer2023adversarial} proposed to introduce adversarial training objective into fine-tuning diffusion models, which dramatically speeds up the sampling process and allows for one-step generation.
They supposed that the Gaussian assumption of diffusion process does not hold anymore when the inference steps are extremely low, and therefore enabling the generative model to output samples in a single forward step by adversarial objective~\cite{xiao2021tackling,xu2023semi}.
We note that the functionality of our coherent reward model $R_\gamma(\cdot)$ is very similar to that of the discriminator in adversarial training, except for the different granularity in prediction.
To this end, unlike the aesthetic and alignment reward models, which necessitate offline learning prior to fine-tuning, the coherent reward model can be learned online and seamlessly integrated into the fine-tuning process. 
The adversarial objective of \textbf{generator} that raises the score of output image is also in compliance with our feedback learning in \cref{eq:reward}, we can simply achieve adversarial training by incorporating the optimization of $R_\gamma(\cdot)$ into fine-tuning to serve as a \textbf{discriminator}.
Thus the \cref{eq:reward} can be reformulated as follows:
\begin{equation}\label{eq:final_reward}
    \mathcal{L}_{reward}(\phi)=-\mathop{\mathbb{E}}\limits_{(x,m,c)\sim\mathcal{D}_{train}\atop t^\prime\sim[1,T_2]}\sum_{\theta\in\{\alpha,\beta,\gamma\}}\log\sigma(R_\theta(c, G_\phi(x,m,c,t^\prime))).
\end{equation}
For completeness, we also rewrite the overall training objective as,
\begin{equation}\label{eq:adversarial}\begin{aligned}
    & \mathcal{L}^G(\phi)=\mathcal{L}_{reward}+\eta(\mathcal{L}_{reg}+\mathcal{L}_{vgg}), \\
    & \mathcal{L}^D(\gamma)=-\mathbb{E}_{(x,m,c)\sim\mathcal{D}_{train}, t^\prime\sim[1,T_2]}[\log\sigma(R_\gamma(z))+\log(1-\sigma(R_\gamma(z^\prime)))].
\end{aligned}\end{equation}


\noindent\textbf{Progressive training.}
To expedite the sampling process, we employ a progressive training strategy where we gradually reduce the optimization time steps $T$. Surprisingly, we find that the quality of the generated images does not significantly degrade under the supervisor of reward models. This approach strikes a fine balance between performance and speed, leading to compelling results.
In our experiments, we adopt a two-phase progressive strategy. During phase 1, we set the optimization time steps as $T=20$, $T_1=15$, and $T_2=10$. In phase 2, we further decrease the time steps to $T=8$, $T_1=6$, and $T_2=3$. Notably, we achieve remarkable outcomes without employing any distillation operations, relying solely on the inheritance of model parameters.


\section{Experiments}

\subsection{Implementation Details}

\subsubsection{Dataset.}
The fine-tuning dataset, denoted as $\mathcal{D}_{train}$, consists of a substantial collection of 7,562,283 images encompassing diverse domains such as real-world scenes, authentic portraits, and computer-generated (CG) images. To enhance the model's generalization ability and generation quality, we adopted a meticulous fine-grained masking strategy inspired by StableDiffusion\cite{rombach2022high}.
Our masking strategy encompasses four distinct types of masks: global masks, irregular shape masks, square masks, and outward expansion masks. Each mask type corresponds to a specific probability value, which is randomly selected and applied to images during the training process. Moreover, we devised a specialized masking strategy tailored for Inpainting-Editing tasks. Leveraging instance-level data, which identifies specific objects within images, we introduced random dilation operations to coarsen the masks during training. These coarsened masks were then integrated with randomly generated masks surrounding the instances. This approach not only enhances the model's performance in instruction-based image editing tasks but also facilitates more accurate instance generation, ultimately leading to superior quality outputs.




To evaluate the performance of our approach, we conducted comprehensive qualitative and quantitative assessments using two distinct datasets.
The first dataset, UserBench, was meticulously curated by gathering a vast amount of user-customized image-text matching data from various online sources. This dataset proved invaluable for evaluating image inpainting and outpainting tasks. From this extensive collection, we judiciously handpicked 100 high-quality image-text matching pairs to serve as our test data. We also leverage the experimental results from this dataset to collect and report human preferences.
The second dataset, EditBench \cite{wang2023imagen}, presents a novel benchmark specifically tailored for text-guided image inpainting. Consisting of 240 images, each image within EditBench is paired with a corresponding mask that precisely delineates the region within the image to be modified through inpainting.

\subsubsection{Training Setting.}
To facilitate the perceptual feedback learning stage, we employed a relatively small learning rate of 2e-06, complemented by a learning rate scheduling strategy that encompassed a warm-up phase consisting of 1000 iterations. Furthermore, to ensure stability in model parameter updates, we incorporated an exponential moving average (EMA) decay parameter set to 0.9999.
Instead of employing 100\% noise as in ReFL\cite{xu2023imagereward}, we introduced a 50
The weight assigned to the perceptual feedback loss was set to 0.01. During the adversarial acceleration stage, we maintained similar settings to the perceptual feedback learning stage, with an additional adversarial loss weighted 0.05.

\subsection{Evaluation Principles and Criteria}


\noindent\textbf{Subjective metrics.}
To assess the robustness of our proposed method, we conducted a comprehensive subjective evaluation involving both expert evaluations and a large number of volunteer participants.
Expert evaluators were tasked with individually assessing each generated image and assigning scores based on three key aspects: \textit{coherence}, \textit{structure}, and \textit{aesthetics}. These aspects were rated on a scale of 1 to 5, with higher scores indicating superior generation quality:
1) \textit{Coherence} focused on evaluating the consistency of texture, style, and color between the generated region and the original image.
2) \textit{Structure} emphasized the clarity, sharpness, and absence of deformations or mutilations, particularly in human body parts.
3) \textit{Aesthetics} gauged the overall level of creativity and diversity exhibited by the generated images.
In addition to expert evaluations, we also sought the opinions of a large number of volunteers, with over 36,000 samples collected. These volunteers were presented with pairs of generated images and asked to choose between ``Good'', ``Same'', or ``Bad'', representing their preference in terms of GSB (Good-Same-Bad).

\noindent\textbf{Objective metrics.}
In this study, we also incorporate objective text-image alignment metrics, specifically CLIPScore~\cite{hessel2021clipscore,radford2021learning} and BLIPScore~\cite{li2022blip}, to comprehensively evaluate the  alignment of our models. 

\subsection{Comparisons with State of the arts}

We compare our method with concurrent state-of-the-art generative image editing systems such as Adobe~\cite{adobe}, Canva~\cite{canva} and MeiTu~\cite{meitu}.
The comparisons cover three different tasks, including outpainting, inpainting-editing and inpainting-erasing.
The inpainting editing will specify the content to be generated for the region of interest in the prompt.
In contrast, inpainting-erasing requires the model to remove content within it and be as consistent as possible. 
Since the erased image has little change, experts were not asked to score aesthetics for user study.

\begin{table}[t]
    \centering
    \scriptsize
    \caption{
    Comparisons with state-of-the-art generative image editing systems in terms of coherence, structure and aesthetic scored by experts.
    More than 6000 image-text pairs are randomly sampled for each task and we report the average scores.
    }
    \vspace{-0.2cm}
    \begin{tabular}{ccccccccc}
        \toprule
        \multirow{2}[2]{*}{\textbf{Method}} & \multicolumn{3}{c}{\textbf{Outpainting}} & \multicolumn{3}{c}{\textbf{Inpainting-Editing}} & \multicolumn{2}{c}{\textbf{Inpainting-Erasing}} \\
        \cmidrule(lr){2-4}\cmidrule(lr){5-7}\cmidrule(lr){8-9}
        & \textit{coherence} & \textit{structure} & \textit{aesthetic} & \textit{coherence} & \textit{structure} & \textit{aesthetic} & \textit{coherence} & \textit{structure} \\
        \midrule
        MeiTu~\cite{meitu} & 3.01 & 2.73 & 2.75 & 2.77 & 2.89 & 2.51 & 3.31 & 3.25 \\
        Canva~\cite{canva} & 2.72 & 2.85 & 2.65 & 3.42 & \underline{3.40} & 3.08 & 2.92 & 2.90 \\
        Adobe~\cite{adobe} & \underline{3.52} & \underline{3.07} & \underline{3.14} & \underline{3.46} & \textbf{3.60} & \underline{3.22} & \underline{3.85} & \textbf{4.28} \\
        \textbf{ByteEdit} & \textbf{3.54} & \textbf{3.25} & \textbf{3.26} & \textbf{3.73} & 3.39 & \textbf{3.25} & \textbf{3.99} & \underline{4.03} \\
        \bottomrule
    \end{tabular}
    \label{tab:user_study}
\end{table}

\begin{figure}[t]
    \centering
    \includegraphics[width=\linewidth]{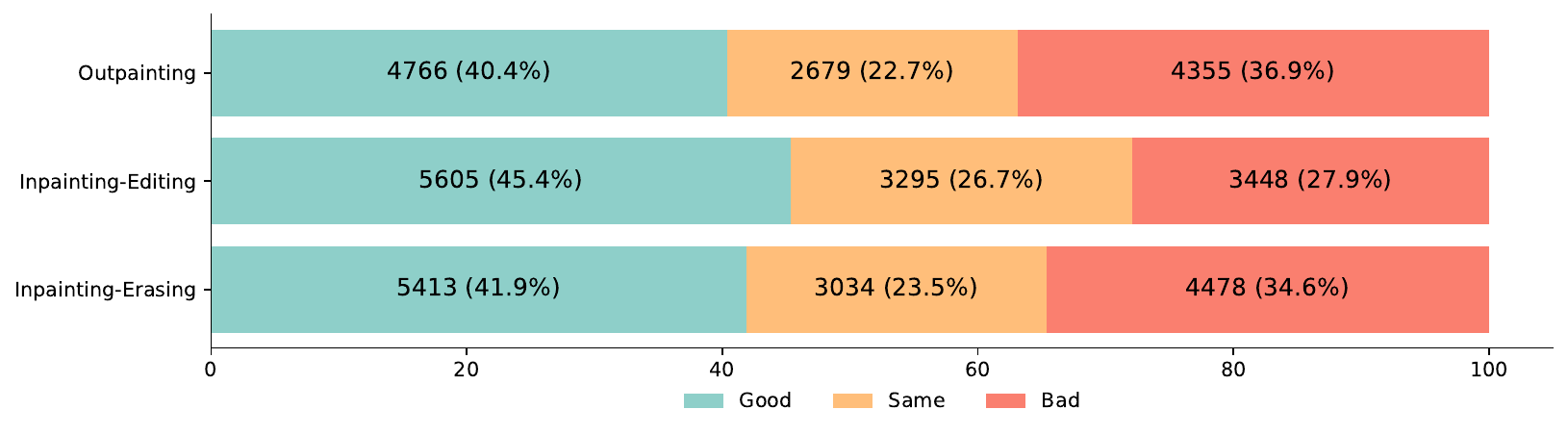}
    \vspace{-0.6cm}
    \caption{
    Comparisons with state-of-the-art generative image editing systems in terms of human preference (i.e. GSB).
    More than 12,000 samples are collected for each task.
    For simplicity and to minimize the difficulty of collecting a large number of user opinions, we only offer the generated images by Adobe and our ByteEdit to the volunteers.
    ``Good'' indicates the generated images by our ByteEdit is preferred and vice versa.
    }
    \vspace{-0.3cm}
    \label{fig:user_study}
\end{figure}

\noindent\textbf{User study.}
The average scores evaluated by experts are shown in \cref{tab:user_study}.
From the results, our ByteEdit significantly outperforms the state-of-the-art across different metrics in the outpainting task.
It demonstrates that our method works well to expand images based on existing content and maintains superior consistency, structural integrity, and creativity.
As for the inpainting tasks, our method also can provide the most coherent edited or erased images.
To further investigate the gap between Adobe and our proposed ByteEdit, we solicited feedback from a large number of volunteers on the images generated by both, and the results are illustrated in \cref{fig:user_study}.
The results show that users generally found the images we generated to be more natural in overall perception.
Our GSB superiority percentages (i.e. (G+S)/(S+B) * 100\%) on three different tasks are 105\%, 163\%, and 112\%, respectively.

\begin{table}[t]
\centering
    \scriptsize
\caption{The quantitative results of ByteEdit and recent state-of-the-art approaches.}
\label{tab:quant_result}
\begin{tabular}{ccccccccc}
\toprule
\multirow{2}[2]{*}{\textbf{Metrics}} & \multicolumn{4}{c}{\textbf{UserBench}} & \multicolumn{4}{c}{\textbf{EditBench}} \\
\cmidrule(lr){2-5}\cmidrule(lr){6-9} & \textit{Meitu}~\cite{meitu} & \textit{Canva}~\cite{canva} & \textit{Adobe}~\cite{adobe} & \textit{\textbf{ByteEdit}} & \textit{DiffEdit}~\cite{couairon2022diffedit} & \textit{BLD}~\cite{avrahami2023blended} & \textit{EMILIE}~\cite{joseph2024iterative} & \textit{\textbf{ByteEdit}} \\
\midrule
CLIPScore & 0.235 & 0.241 & 0.237 & \textbf{0.255} & 0.272 & 0.280 & 0.311 & \textbf{0.329} \\
BLIPScore & 0.174 & 0.467 & 0.450 & \textbf{0.687} & 0.582 & 0.596 & 0.620 & \textbf{0.691} \\
\bottomrule
\end{tabular}
\end{table}

\begin{figure}[t]
    \centering
    \includegraphics[width=1.0\linewidth]{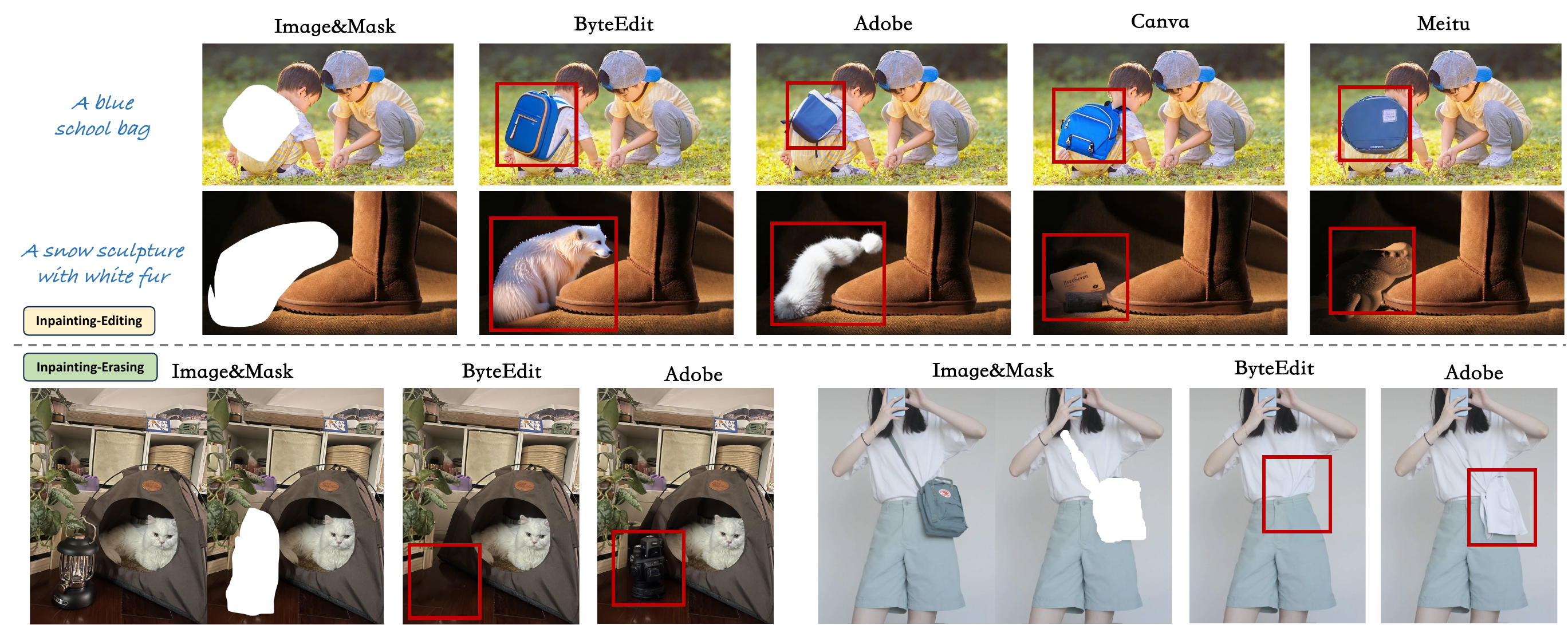}
    \caption{
  Qualitative comparison in inpainting. We highlight key areas with red boxes.
    }
    \label{fig:inpaint_sota}
\end{figure}

\begin{figure}[t]
    \centering
    \includegraphics[width=1.0\linewidth]{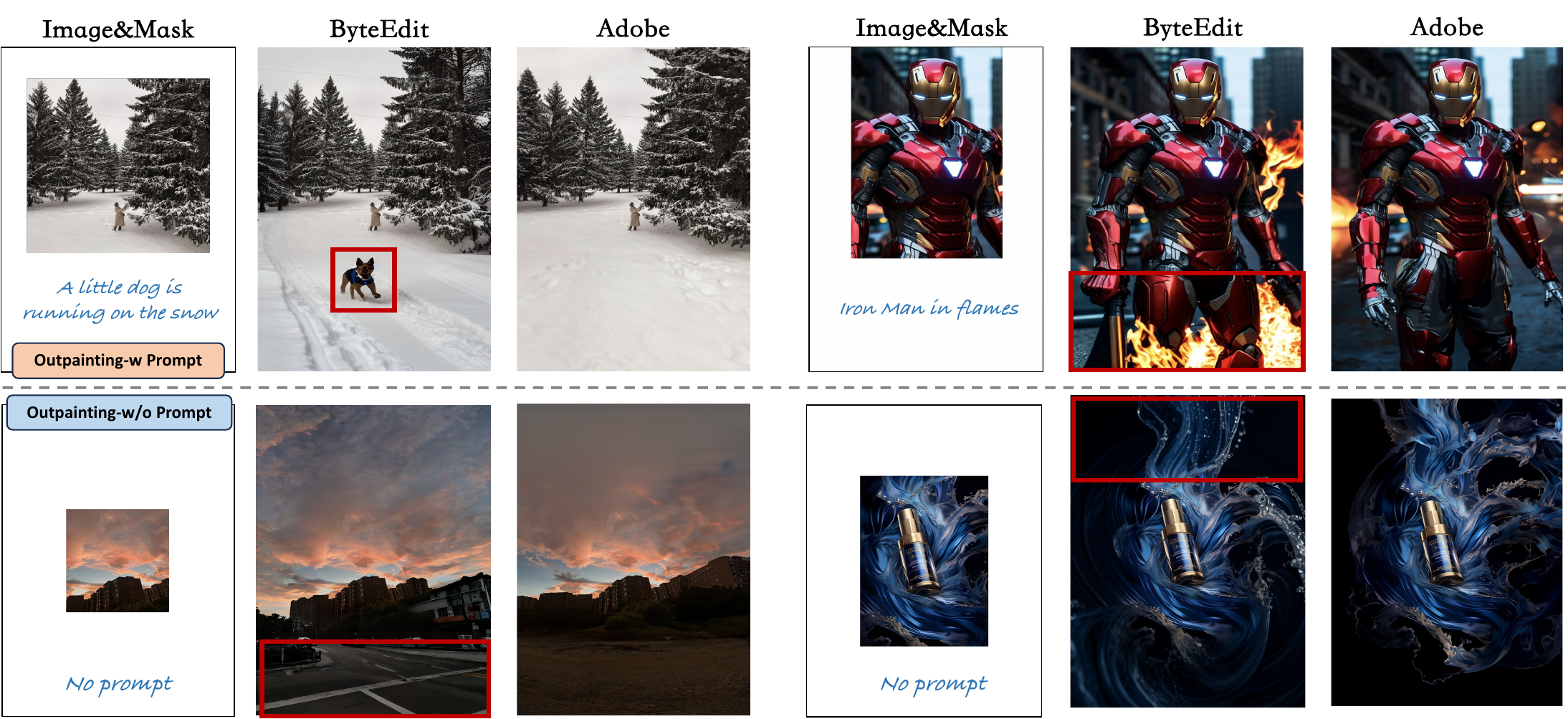}
    \caption{
  Qualitative comparison in outpainting. We highlight key areas with red boxes.
    }
    \label{fig:outpaint_sota}
\end{figure}

\noindent\textbf{Quantitative comparison.}
To quantitatively evaluate the performance of our method compared to other approaches,  we conduct a quantitative evaluation of how well the edited image can capture the semantics of the edit instruction successfully by measuring the CLIPScore and BLIPScore. 
We conduct the experiment in inpainting-editing task and the results are provided in Table \ref{tab:quant_result}.
From the UserBench against state-of-the-art generative image editing systems, we noticed that the score results are not completely consistent with user opinion. 
Nevertheless, our method is still ahead of the second-place Canva by \textbf{5.8\%}(+0.014) and \textbf{47.1\%}(+0.22) in terms of CLIPScore and BLIPScore, respectively.
As for the EditBench, we follow \cite{joseph2024iterative} to compare our method with several concurrent editing approaches, i.e. DiffEdit~\cite{couairon2022diffedit}, BLD~\cite{avrahami2023blended} and EMILIE~\cite{joseph2024iterative}.
It demonstrates that the ByteEdit consistently yields the state-of-the-art performance, which shows our superior quality, consistency and instruction adherence.

\noindent\textbf{Qualitative comparison.}
In Figure \ref{fig:inpaint_sota} and \ref{fig:outpaint_sota}, we visualize samples produced by different systems under different tasks.
It clearly shows that our method exhibits a superior performance for learning both coherence and aesthetic.
For the inpainting task, the ByteEdit consistently follows the user-specified instructions and generates coherent images with better image-text alignment.
It is worth noting that our system allows both prompt and prompt-free generation when it comes to outpainting, which has broader application scenarios in reality.


\subsection{Ablation Studies}

\begin{figure}[t]
    \centering
    \subfloat[w/ vs w/o PeFL]{\includegraphics[width=0.5\linewidth]{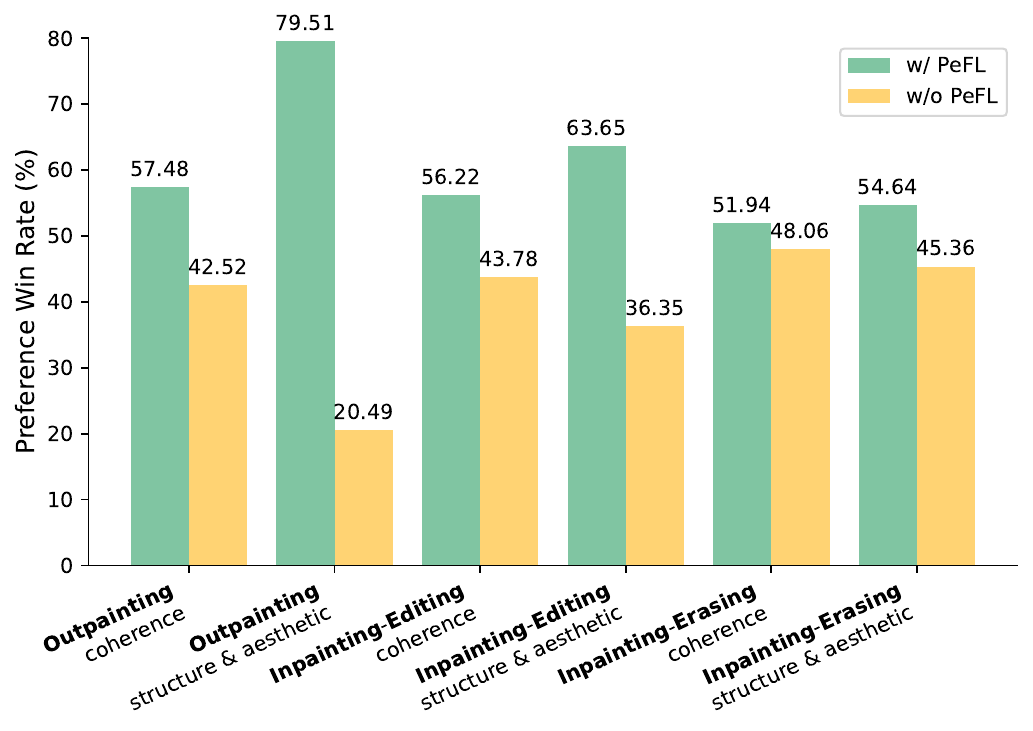}}
    \subfloat[w/ vs w/o Acceleration]{\includegraphics[width=0.5\linewidth]{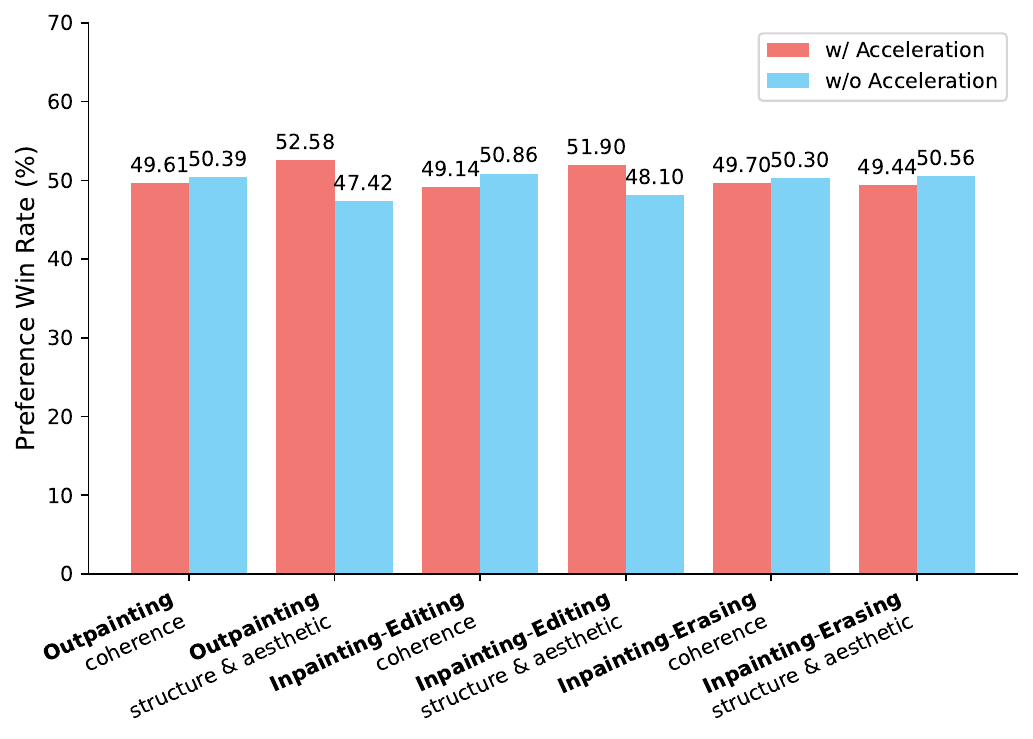}}
    \caption{
    Human Perference Evaluation on our proposed PeFL and Acceleration.
    }
    \label{fig:ablation}
\end{figure}

\begin{figure}[t]
    \centering
    \includegraphics[width=1.0\linewidth]{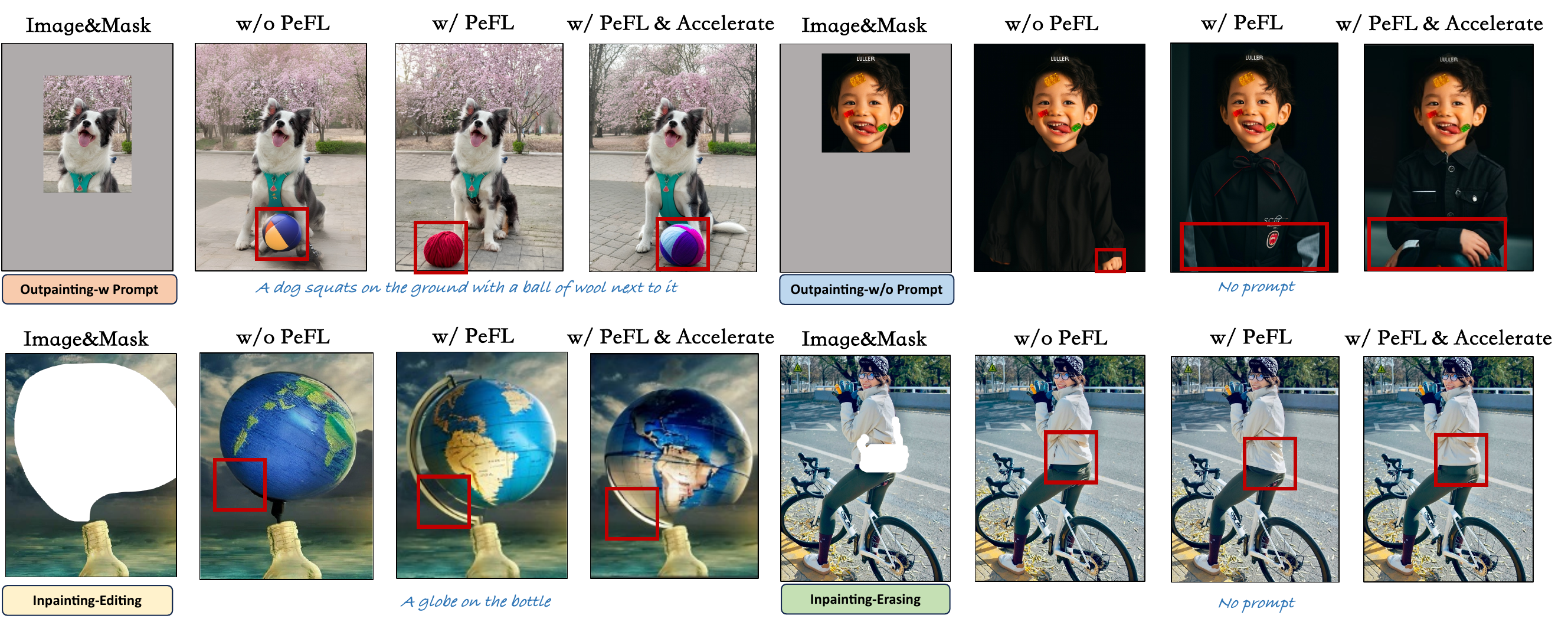}
    \caption{
  Ablation Studies Visualization.
    }
    \label{fig:ablation_vis}
\end{figure}

In Figure \ref{fig:ablation}, we conduct ablation experiments on both our proposed PeFL and acceleration strategy.
The experts were asked to choose GSB preference and we report the human preference rates in the figure, which are calculated as (G+S)/[(G+S)+(S+B)] * 100\% for win and (S+B)/[(G+S)+(S+B)] * 100\% for lose, respectively.
The evaluation is similar to the user study, except that we combine structure and aesthetics to reduce evaluation costs.
More visualizations are also included in Figure \ref{fig:ablation_vis}.

\noindent\textbf{PeFL preference.}
From the results in Figure \ref{fig:ablation}(a), our proposed PeFL significantly improves the generation quality, outperforming the baseline on all different tasks.
Especially in the outpainting task with PeFL, our method exceeds the baseline by about 60\% in terms of structure and aesthetic, which is consistent with the edited image shown at the top of Figure \ref{fig:ablation_vis} that is more realistic and conforms to the rules of realistic physics.

\noindent\textbf{Acceleration preference.}
In Figure \ref{fig:ablation}(b), we demonstrate that our model has no significant loss in either consistency or structure and aesthetic with the progressive training strategy.
To our surprise, we have even achieved both increasing speed and quality in the outpainting and inpainting-editing tasks.
Based on our experimental observations,  this phenomenon can be attributed to two underlying factors:
i) Stable Training: By considering the discriminator as a reward model, trainable reward model offers flexible and robust supervision for PeFL, alleviating issues related to model over-optimization; ii) Expand Supervision Scope: The incorporation of adversarial supervision enables us to extend the time step of PEFL optimization. Consequently, even at high-noise stages, such as step 999, the model can still benefit from PeFL supervision, further driving improvements in model performance.
The visualization at the bottom of Figure \ref{fig:ablation_vis} also verifies this, where the outputs become more realistic and natural after acceleration.
\section{Discussion}

ByteEdit has demonstrated remarkable performance across various editing tasks. However, several promising directions warrant further exploration:

\begin{itemize}
\item \textit{Performance}: One avenue for improvement lies in developing more targeted reward models tailored to specific editing tasks. By refining the reward models, we can potentially unlock even higher levels of performance and generate more precise and desirable output.

\item \textit{Acceleration}: Another area of interest is investigating how ByteEdit can be further integrated with advanced techniques such as LCM and SDXL-turbo to achieve accelerated processing speeds. 

\item \textit{Task}: Expanding the capabilities of ByteEdit beyond image editing to domains like video editing or instruction editing holds significant potential. 

\end{itemize}
By incorporating human feedback to optimize generative image editing, ByteEdit can greatly enhance the practicality and usability in real-world scenarios. We hope that our work will provide valuable insights and inspire deeper reflections in this field, propelling its future development.


\section{Conclusion}
ByteEdit is a feedback learning framework that designs to enhance generation quality, consistency, and inference speed in diffusion-based generative image editing tasks. Through extensive user evaluations, we demonstrate that ByteEdit surpasses leading generative image editing products. And its effectiveness and superior performance position ByteEdit as a state-of-the-art tool in the field.


%
%
\bibliographystyle{splncs04}
\bibliography{main}

\begin{thebibliography}{10}
\providecommand{\url}[1]{\texttt{#1}}
\providecommand{\urlprefix}{URL }
\providecommand{\doi}[1]{https://doi.org/#1}

\bibitem{adobe}
Adobe firefly - free generative ai for creatives. \url{https://www.adobe.com/products/firefly.html}

\bibitem{canva}
Free ai image generator: Online text to image app | canva. \url{https://www.canva.com/ai-image-generator/}

\bibitem{meitu}
Miraclevision. \url{https://ai.meitu.com/index/}

\bibitem{avrahami2023blended}
Avrahami, O., Fried, O., Lischinski, D.: Blended latent diffusion. ACM Transactions on Graphics (TOG)  \textbf{42}(4),  1--11 (2023)

\bibitem{avrahami2022blended}
Avrahami, O., Lischinski, D., Fried, O.: Blended diffusion for text-driven editing of natural images. In: Proceedings of the IEEE/CVF Conference on Computer Vision and Pattern Recognition. pp. 18208--18218 (2022)

\bibitem{chen2023anydoor}
Chen, X., Huang, L., Liu, Y., Shen, Y., Zhao, D., Zhao, H.: Anydoor: Zero-shot object-level image customization. arXiv preprint arXiv:2307.09481  (2023)

\bibitem{chen2015microsoft}
Chen, X., Fang, H., Lin, T.Y., Vedantam, R., Gupta, S., Doll{\'a}r, P., Zitnick, C.L.: Microsoft coco captions: Data collection and evaluation server. arXiv preprint arXiv:1504.00325  (2015)

\bibitem{couairon2022diffedit}
Couairon, G., Verbeek, J., Schwenk, H., Cord, M.: Diffedit: Diffusion-based semantic image editing with mask guidance. arXiv preprint arXiv:2210.11427  (2022)

\bibitem{dong2023raft}
Dong, H., Xiong, W., Goyal, D., Pan, R., Diao, S., Zhang, J., Shum, K., Zhang, T.: Raft: Reward ranked finetuning for generative foundation model alignment. arXiv preprint arXiv:2304.06767  (2023)

\bibitem{dosovitskiy2020image}
Dosovitskiy, A., Beyer, L., Kolesnikov, A., Weissenborn, D., Zhai, X., Unterthiner, T., Dehghani, M., Minderer, M., Heigold, G., Gelly, S., et~al.: An image is worth 16x16 words: Transformers for image recognition at scale. arXiv preprint arXiv:2010.11929  (2020)

\bibitem{hessel2021clipscore}
Hessel, J., Holtzman, A., Forbes, M., Bras, R.L., Choi, Y.: Clipscore: A reference-free evaluation metric for image captioning. arXiv preprint arXiv:2104.08718  (2021)

\bibitem{ho2020denoising}
Ho, J., Jain, A., Abbeel, P.: Denoising diffusion probabilistic models. Advances in neural information processing systems  \textbf{33},  6840--6851 (2020)

\bibitem{isajanyan2024social}
Isajanyan, A., Shatveryan, A., Kocharyan, D., Wang, Z., Shi, H.: Social reward: Evaluating and enhancing generative ai through million-user feedback from an online creative community. arXiv preprint arXiv:2402.09872  (2024)

\bibitem{joseph2024iterative}
Joseph, K., Udhayanan, P., Shukla, T., Agarwal, A., Karanam, S., Goswami, K., Srinivasan, B.V.: Iterative multi-granular image editing using diffusion models. In: Proceedings of the IEEE/CVF Winter Conference on Applications of Computer Vision. pp. 8107--8116 (2024)

\bibitem{kirillov2023segment}
Kirillov, A., Mintun, E., Ravi, N., Mao, H., Rolland, C., Gustafson, L., Xiao, T., Whitehead, S., Berg, A.C., Lo, W.Y., et~al.: Segment anything. arXiv preprint arXiv:2304.02643  (2023)

\bibitem{kirstain2024pick}
Kirstain, Y., Polyak, A., Singer, U., Matiana, S., Penna, J., Levy, O.: Pick-a-pic: An open dataset of user preferences for text-to-image generation. Advances in Neural Information Processing Systems  \textbf{36} (2024)

\bibitem{lee2023aligning}
Lee, K., Liu, H., Ryu, M., Watkins, O., Du, Y., Boutilier, C., Abbeel, P., Ghavamzadeh, M., Gu, S.S.: Aligning text-to-image models using human feedback. arXiv preprint arXiv:2302.12192  (2023)

\bibitem{li2022blip}
Li, J., Li, D., Xiong, C., Hoi, S.: Blip: Bootstrapping language-image pre-training for unified vision-language understanding and generation. In: International Conference on Machine Learning. pp. 12888--12900. PMLR (2022)

\bibitem{liu2023visual}
Liu, H., Li, C., Wu, Q., Lee, Y.J.: Visual instruction tuning. arXiv preprint arXiv:2304.08485  (2023)

\bibitem{lu2024coarse}
Lu, Y., Zhang, M., Ma, A.J., Xie, X., Lai, J.H.: Coarse-to-fine latent diffusion for pose-guided person image synthesis. arXiv preprint arXiv:2402.18078  (2024)

\bibitem{lugmayr2022repaint}
Lugmayr, A., Danelljan, M., Romero, A., Yu, F., Timofte, R., Van~Gool, L.: Repaint: Inpainting using denoising diffusion probabilistic models. In: Proceedings of the IEEE/CVF Conference on Computer Vision and Pattern Recognition. pp. 11461--11471 (2022)

\bibitem{van2008visualizing}
Van~der Maaten, L., Hinton, G.: Visualizing data using t-sne. Journal of machine learning research  \textbf{9}(11) (2008)

\bibitem{nichol2021glide}
Nichol, A., Dhariwal, P., Ramesh, A., Shyam, P., Mishkin, P., McGrew, B., Sutskever, I., Chen, M.: Glide: Towards photorealistic image generation and editing with text-guided diffusion models. arXiv preprint arXiv:2112.10741  (2021)

\bibitem{podell2023sdxl}
Podell, D., English, Z., Lacey, K., Blattmann, A., Dockhorn, T., M{\"u}ller, J., Penna, J., Rombach, R.: Sdxl: Improving latent diffusion models for high-resolution image synthesis. arXiv preprint arXiv:2307.01952  (2023)

\bibitem{qin2024diffusiongpt}
Qin, J., Wu, J., Chen, W., Ren, Y., Li, H., Wu, H., Xiao, X., Wang, R., Wen, S.: Diffusiongpt: Llm-driven text-to-image generation system. arXiv preprint arXiv:2401.10061  (2024)

\bibitem{radford2021learning}
Radford, A., Kim, J.W., Hallacy, C., Ramesh, A., Goh, G., Agarwal, S., Sastry, G., Askell, A., Mishkin, P., Clark, J., et~al.: Learning transferable visual models from natural language supervision. In: International conference on machine learning. pp. 8748--8763. PMLR (2021)

\bibitem{ren2023ugc}
Ren, Y., Wu, J., Zhang, P., Zhang, M., Xiao, X., He, Q., Wang, R., Zheng, M., Pan, X.: Ugc: Unified gan compression for efficient image-to-image translation. In: Proceedings of the IEEE/CVF International Conference on Computer Vision. pp. 17281--17291 (2023)

\bibitem{rombach2022high}
Rombach, R., Blattmann, A., Lorenz, D., Esser, P., Ommer, B.: High-resolution image synthesis with latent diffusion models. In: Proceedings of the IEEE/CVF conference on computer vision and pattern recognition. pp. 10684--10695 (2022)

\bibitem{ronneberger2015u}
Ronneberger, O., Fischer, P., Brox, T.: U-net: Convolutional networks for biomedical image segmentation. In: Medical Image Computing and Computer-Assisted Intervention--MICCAI 2015: 18th International Conference, Munich, Germany, October 5-9, 2015, Proceedings, Part III 18. pp. 234--241. Springer (2015)

\bibitem{sauer2023adversarial}
Sauer, A., Lorenz, D., Blattmann, A., Rombach, R.: Adversarial diffusion distillation. arXiv preprint arXiv:2311.17042  (2023)

\bibitem{schuhmann2022laion}
Schuhmann, C., Beaumont, R., Vencu, R., Gordon, C., Wightman, R., Cherti, M., Coombes, T., Katta, A., Mullis, C., Wortsman, M., et~al.: Laion-5b: An open large-scale dataset for training next generation image-text models. Advances in Neural Information Processing Systems  \textbf{35},  25278--25294 (2022)

\bibitem{simonyan2014very}
Simonyan, K., Zisserman, A.: Very deep convolutional networks for large-scale image recognition. arXiv preprint arXiv:1409.1556  (2014)

\bibitem{turc2022midjourney}
Turc, I., Nemade, G.: Midjourney user prompts \& generated images (250k) (2022). \doi{10.34740/KAGGLE/DS/2349267}

\bibitem{wang2023imagen}
Wang, S., Saharia, C., Montgomery, C., Pont-Tuset, J., Noy, S., Pellegrini, S., Onoe, Y., Laszlo, S., Fleet, D.J., Soricut, R., et~al.: Imagen editor and editbench: Advancing and evaluating text-guided image inpainting. In: Proceedings of the IEEE/CVF Conference on Computer Vision and Pattern Recognition. pp. 18359--18369 (2023)

\bibitem{wu2023human}
Wu, X., Sun, K., Zhu, F., Zhao, R., Li, H.: Human preference score: Better aligning text-to-image models with human preference. In: Proceedings of the IEEE/CVF International Conference on Computer Vision. pp. 2096--2105 (2023)

\bibitem{xiao2021tackling}
Xiao, Z., Kreis, K., Vahdat, A.: Tackling the generative learning trilemma with denoising diffusion gans. arXiv preprint arXiv:2112.07804  (2021)

\bibitem{xie2023smartbrush}
Xie, S., Zhang, Z., Lin, Z., Hinz, T., Zhang, K.: Smartbrush: Text and shape guided object inpainting with diffusion model. In: Proceedings of the IEEE/CVF Conference on Computer Vision and Pattern Recognition. pp. 22428--22437 (2023)

\bibitem{xie2023dreaminpainter}
Xie, S., Zhao, Y., Xiao, Z., Chan, K.C., Li, Y., Xu, Y., Zhang, K., Hou, T.: Dreaminpainter: Text-guided subject-driven image inpainting with diffusion models. arXiv preprint arXiv:2312.03771  (2023)

\bibitem{xu2023imagereward}
Xu, J., Liu, X., Wu, Y., Tong, Y., Li, Q., Ding, M., Tang, J., Dong, Y.: Imagereward: Learning and evaluating human preferences for text-to-image generation. arXiv preprint arXiv:2304.05977  (2023)

\bibitem{xu2023semi}
Xu, Y., Gong, M., Xie, S., Wei, W., Grundmann, M., Hou, T., et~al.: Semi-implicit denoising diffusion models (siddms). arXiv preprint arXiv:2306.12511  (2023)

\bibitem{xu2023ufogen}
Xu, Y., Zhao, Y., Xiao, Z., Hou, T.: Ufogen: You forward once large scale text-to-image generation via diffusion gans. arXiv preprint arXiv:2311.09257  (2023)

\bibitem{yang2023paint}
Yang, B., Gu, S., Zhang, B., Zhang, T., Chen, X., Sun, X., Chen, D., Wen, F.: Paint by example: Exemplar-based image editing with diffusion models. In: Proceedings of the IEEE/CVF Conference on Computer Vision and Pattern Recognition. pp. 18381--18391 (2023)

\bibitem{yang2024dense}
Yang, S., Chen, T., Zhou, M.: A dense reward view on aligning text-to-image diffusion with preference. arXiv preprint arXiv:2402.08265  (2024)

\bibitem{yildirim2023inst}
Yildirim, A.B., Baday, V., Erdem, E., Erdem, A., Dundar, A.: Inst-inpaint: Instructing to remove objects with diffusion models. arXiv preprint arXiv:2304.03246  (2023)

\bibitem{yu2023inpaint}
Yu, T., Feng, R., Feng, R., Liu, J., Jin, X., Zeng, W., Chen, Z.: Inpaint anything: Segment anything meets image inpainting. arXiv preprint arXiv:2304.06790  (2023)

\bibitem{yuan2024self}
Yuan, H., Chen, Z., Ji, K., Gu, Q.: Self-play fine-tuning of diffusion models for text-to-image generation. arXiv preprint arXiv:2402.10210  (2024)

\bibitem{zhang2023diffusionengine}
Zhang, M., Wu, J., Ren, Y., Li, M., Qin, J., Xiao, X., Liu, W., Wang, R., Zheng, M., Ma, A.J.: Diffusionengine: Diffusion model is scalable data engine for object detection. arXiv preprint arXiv:2309.03893  (2023)

\bibitem{zhang2024confronting}
Zhang, Z., Zhang, S., Zhan, Y., Luo, Y., Wen, Y., Tao, D.: Confronting reward overoptimization for diffusion models: A perspective of inductive and primacy biases. arXiv preprint arXiv:2402.08552  (2024)

\end{thebibliography}
\end{document}